\documentclass{article}
\usepackage{spconf,amsmath,graphicx}

\title{Automatic Liver Lesion Detection using Cascaded Deep Residual Networks}
\name{Lei Bi, Jinman Kim, Ashnil Kumar, Dagan Feng}
\address{School of Information Technologies, University of Sydney, Australia\\
Med-X Research Institute, Shanghai Jiao Tong University, China
}
\usepackage{graphicx}
\usepackage{marvosym}
\usepackage{hyperref}
\usepackage{enumitem}
\usepackage{amsmath,amssymb,latexsym} 
\usepackage{multirow}
\usepackage{lipsum}
\usepackage{pifont}
\begin{document}
%
\maketitle
\begin{abstract}
Automatic segmentation of liver lesions is a fundamental requirement towards the creation of computer aided diagnosis (CAD) and decision support systems (CDS). Traditional segmentation approaches depend heavily upon hand-crafted features and a priori knowledge of the user. As such, these methods are difficult to adopt within a clinical environment. Recently, deep learning methods based on fully convolutional networks (FCNs) have been successful in many segmentation problems primarily because they leverage a large labelled dataset to hierarchically learn the features that best correspond to the shallow visual appearance as well as the deep semantics of the areas to be segmented. However, FCNs based on a 16 layer VGGNet architecture have limited capacity to add additional layers. Therefore, it is challenging to learn more discriminative features among different classes for FCNs. In this study, we overcome these limitations using deep residual networks (ResNet) to segment liver lesions. ResNet contain skip connections between convolutional layers, which solved the problem of the training degradation of training accuracy in very deep networks and thereby enables the use of additional layers for learning more discriminative features. In addition, we achieve more precise boundary definitions through a novel cascaded ResNet architecture with multi-scale fusion to gradually learn and infer the boundaries of both the liver and the liver lesions. Our proposed method achieved 4\textsuperscript{th} place in the ISBI 2017 Liver Tumor Segmentation Challenge by the submission deadline.
\end{abstract}
\section{Introduction}
\label{sec:intro}

Primary liver cancer is globally considered as the second most common cause of cancer death and is the sixth most frequent cancer in the world \cite{C1}. Computed tomography (CT) is the most commonly used modality for liver lesion evaluation and staging \cite{C2}. Manual measurement of the size of each liver lesion is the norm in routine clinical practice. However, manual liver lesion segmentation is subjective, operator dependent, poorly reproducible and time-consuming. For these reasons, there has been increasing research interest in the development of fully automated liver lesion segmentation methods.

\par

Traditional segmentation methods \cite{C3,C4} depend heavily upon hand-crafted features and a priori knowledge of the user who will refine the segmentation. In addition, these methods usually rely on using low-level features, such as local texture that do not capture image-wide variation. Furthermore, their performance relies on correctly tuning a large number of parameters. Consequently, these methods are highly unreliable for accurate segmentation.

\par

Deep learning methods based on fully convolutional networks (FCN) have recently demonstrated many successes in segmentation problems \cite{C5,C6,C7,C8,C21,C22,C23}. This is primarily attributed to the ability of FCN to leverage large datasets to learn a feature representation that combines low-level appearance information in lower layers with high-level semantic information in the deeper layers \cite{C5}. In addition, FCN can be trained in an end-to-end manner for efficient inference, i.e., images are taken as inputs and the segmentation results are directly output.

\par

However, traditional FCN is based on the VGGNet \cite{C9} architecture, which only has 16 layers for training and thus has limited capacity to learn the discriminative features among different classes. In addition, it has been demonstrated that stacking extra layers results in higher training and validation errors beyond certain depths. Therefore, it is challenging to optimize very deep networks with many layers. In addition, FCN has large receptive fields in the convolutional filters and hence produces coarse outputs at the lesion boundaries. The outputs also lack smoothness, i.e., the labels of similar neighboring pixels may not agree, and therefore producing a segmentation probability maps with an inconsistent spatial appearance. Deep residual net-works (ResNet) with 50, 100, 150 or 1000 layers have achieved state-of-the-art results in image classification and detection problems \cite{C10}. ResNet architectures con-sists of a number of residual blocks that bypasses (skips or shortcuts) a few convolution layers at a time \cite{C10}. The outputs of the shortcut connections are aggregated with the output of the convolution layers, which overcome the limitation of adding extra layers by reducing the training degradation often witnessed in very deep networks. In addition, ResNet can be considered as an ensembles of many shallow networks \cite{C12,C13,C14}, where different networks are connected via these shortcuts and therefore, optimal results can be achieved by averaging the output of the different networks. 

\par

 In the ISBI 2017 Liver Tumor Segmentation Challenge\footnote{https://competitions.codalab.org/competitions/15595} (LiTS), we exploit the deep residual networks for robust liver lesion segmentation and we introduce the following contributions:

\begin{enumerate}[label={(\arabic*)}]

\item

An automatic liver and liver tumor detection method based on deep residual networks that, when compared with traditional VGGNet based FCN architecture, improves both the liver and liver lesion detection accuracy. 

\item

A cascaded ResNet architecture to iteratively refine and constrain the lesion boundaries at both training and testing time. During training, the cascaded ResNet learns from the training data and the estimated results derived from the previous iteration. The ability to learn from the previous iteration optimizes the learning of both the liver and liver lesion boundaries, which are usually difficult to segment. During testing (prediction), the cascaded ResNet uses test (input) images and the estimated probability map derived from previous itera-tions to gradually improve the segmentation accuracy.

\end{enumerate}

\section{Methods}
\label{sec:method}

\subsection{Materials}
\label{sssec:materials}

The LiTS dataset comprises 201 contrast-enhanced abdomen CT studies acquired from 6 medical centers around the world; there were 131 training and 70 test images. Both liver and liver lesion masks (ground truth) were provided in the training data. All the ground truth annotations were carefully prepared under supervision of expert radiologists. We further split the training dataset into 118 studies as for training and 13 studies for validation.

\subsection{Pre-processing}
\label{sssec:Pre-processing}

We set the Hounsfield Unit (HU) value range to [-160, 240] to exclude irrelevant organs and objects; the range was set based on the liver window [-62, 238] given by Sahi et al. \cite{C11} less 100 HU value on the lower bound to ensure all the liver lesions would be captured (Fig. \ref{fig:3}). After HU value adjustment, the voxel values of each 3D volume was normalized into the range [0, 1].

\subsection{Deep Residual Networks for Segmentation}
\label{sssec:Deep Residual Networks for Segmentation}

\begin{figure}
	\begin{center}
			\includegraphics[width=0.5\columnwidth]{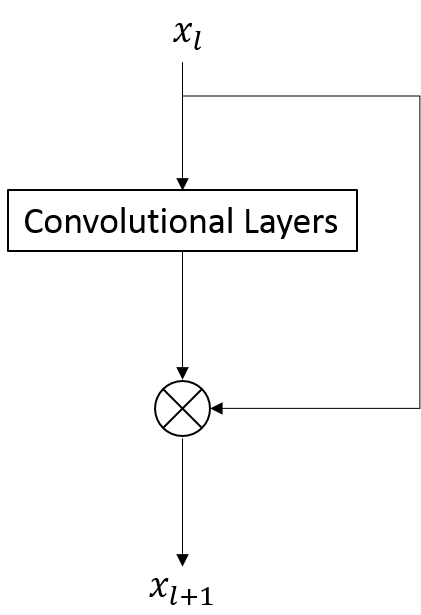}
	\end{center}
	\caption{A sample residual block.}
	\label{fig:1}
\end{figure}

The ResNet architecture consists of a number of residual blocks with each block comprising of several convolution layers, batch normalization layers, and ReLU layers. The residual block consists of several convolutional layers and a skip/shortcut connection to bypass these layers \cite{C10}. The ResNet architecture is able to train deeper networks without training degradation and provides better discriminative features. The residual block is calculated as:

\begin{equation}
{x_{l + 1}} = {x_l} + {\cal F}\left( {{x_l},{\omega _l}} \right)
\end{equation}

where $x_l$ is the input of the $l$-th block in the network and $x_{l+1}$ is its output, ${\cal F}\left(  \cdot  \right)$ is the residual function, and $\omega _l$ are the weight parameters for that block. A sample residual block is shown in Fig. \ref{fig:1}.

We transformed the ResNet architecture into a segmentation model by adding convolutional and deconvolutional layers to upsample the output features maps (as suggested by the FCN architecture of Long et al. \cite{C5}),  and dilating the feature maps derived from ResNet to create the score mask \cite{C13,C15}. Our architecture is shown in Table \ref{table:tab1}.

\begin{table*}[t]
\centering
\caption{Deep residual network architecture for segmentation.}
\label{table:tab1}
\def\arraystretch{1.2}
\begin{tabular}{llll}
\hline
\textbf{Output}  & \textbf{Convolution Type}              & \textbf{Residual Block} & \textbf{Number} \\ \hline
504$\times$504 & 3$\times$3, 64, stride 1             & \ding{56}               & 1      \\ \hline
252$\times$252 & 3$\times$3, 128, stride 2            & \ding{52}              & 1      \\
        & 3$\times$3, 128, stride 1            &                &        \\ \hline
252$\times$252 & 3$\times$3, 128, stride 1            & \ding{52}              & 2      \\
        & 3$\times$3, 128, stride 1            &                &        \\ \hline
126$\times$126 & 3$\times$3, 256, stride 2            & \ding{52}              & 1      \\
        & 3$\times$3, 256, stride 1            &                &        \\ \hline
126$\times$126 & 3$\times$3, 256, stride 1            & \ding{52}              & 2      \\
        & 3$\times$3, 256, stride 1            &                &        \\ \hline
63$\times$63   & 3$\times$3, 512, stride 2            & \ding{52}              & 1      \\
        & 3$\times$3, 512, stride 1            &                &        \\ \hline
63$\times$63   & 3$\times$3, 512, stride 1            & \ding{52}              & 5      \\
        & 3$\times$3, 512, stride 1            &                &        \\ \hline
63$\times$63   & 3$\times$3, 512, stride 1            & \ding{52}              & 1      \\
        & 3$\times$3, 1024, stride 1, dilate 2 &                &        \\ \hline
63$\times$63   & 3$\times$3, 512, stride 1, dilate 2  & \ding{52}              & 2      \\
        & 3$\times$3, 1024, stride 1, dilate 2 &                &        \\ \hline
63$\times$63   & 1$\times$1, 512, stride 1            & \ding{52}              & 1      \\
        & 3$\times$3, 1024, stride 1, dilate 4 &                &        \\
        & 1$\times$1, 2048, stride 1           &                &        \\ \hline
63$\times$63   & 1$\times$1, 1024, stride 1           & \ding{52}              & 1      \\
        & 3$\times$3, 2048, stride 1, dilate 4 &                &        \\
        & 1$\times$1, 4096, stride 1           &                &        \\ \hline
63$\times$63   & 3$\times$3, 512, stride 1, dilate 12 & \ding{56}              & 1      \\ \hline
63$\times$63   & 3$\times$3, 2, stride 1, dilate 12   & \ding{56}              & 1      \\ \hline
\end{tabular}
\end{table*}

\subsection{Cascaded Deep Residual Networks for Segmentation}
\label{sssec:Cascaded Deep Residual Networks for Segmentation}

The whole deep residual network for segmentation can be defined as:
\begin{equation}
Y = {F_S}\left( {I;w} \right)
\end{equation}

where $Y$ is the output prediction, $I$ is the input image, $F_S$ denotes the feature map produced by the stacked convolutional layers (or residual block) with a list of stride or dilation values $S$, and $w$ denotes the learned parameters. 

\begin{figure*}
	\begin{center}
			\includegraphics[width=0.8\textwidth]{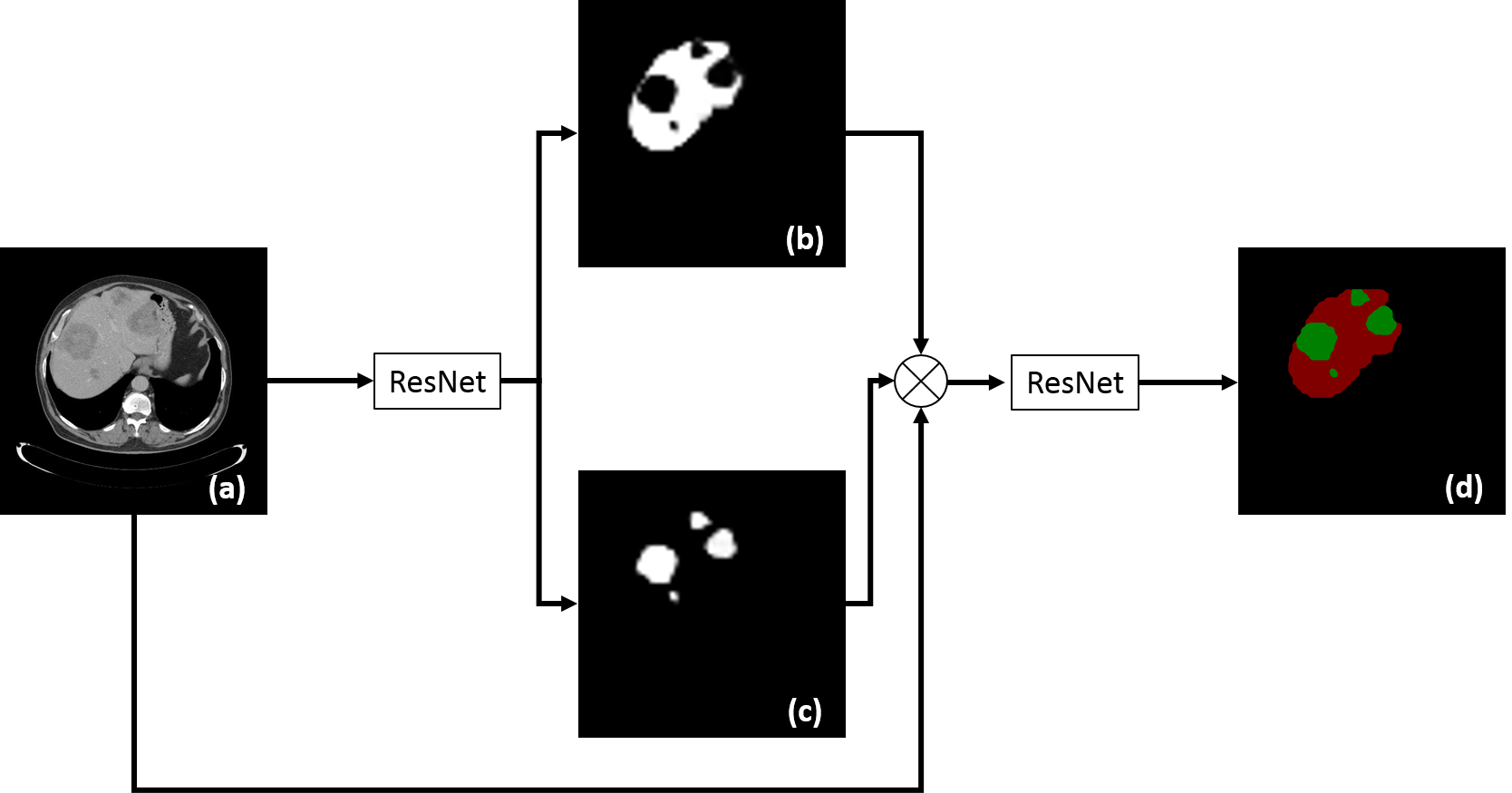}
	\end{center}
	\caption{Overview of the proposed cascaded ResNet architecture, where (a) is the input CT image; (b) and (c) are the produced probability map of the liver and the liver lesions, respectively; and (c) is the final predication from the cascaded ResNet for the liver and the liver lesions.}
	\label{fig:2}
\end{figure*}

Our cascaded ResNet embeds the probability map produced at the previous deep residual networks for training and testing (as exemplified in Fig. \ref{fig:2}) and the calculation can be defined as:

\begin{equation}
P = {F_S}\left( {I,{Y^t},{Y^v};w} \right)
\end{equation}

where $P$ is the output of the prediction of the cascaded ResNet, $Y^t$ and $Y^v$ denote the probability map derived from ResNet for predicting the tumor and liver regions, respectively. During testing, a multi-scale integration approach was used, where we resized the image into a number of scales (size from 512$\times$512 to 640$\times$640 with an increment of 32). The final output was produced by averaging the multi-scale outputs. For post-processing, a morphological filter was used to fill the holes for individual axial slices; no other post-processing was used.

\subsection{Implementation Details}
\label{sssec:Implementation Details}

Our training and segmentation were performed on 2D axial slices and this is attributed to the fact that within many studies, there are duplicated slices  (e.g., liver was scanned twice in a single CT study). Furthermore, it will be too time-consuming to train all the individual slices ($\sim$60,000 slices in total, would take more than 1 month to train all slices).  Therefore, we randomly selected 8,802 slices, consisting of 4,401 slices presenting both liver and liver lesions, and another 4,401 slices where both liver and liver lesions were not present.  

\par

The training process can be defined as minimizing the overall per-pixel loss, with iterative updates of the networks’ weight parameters using stochastic gradient descent (SGD). Research has suggested that fine-tuning can improve the robustness of the trained model, where the lower layers of the fine-tuned network are more general filters (trained on general images) while those in the higher layers are more specific to the target problem. Therefore, we trained the proposed cascaded ResNet via fine-tuning; we first fine-tuned the pre-trained model trained on ImageNet dataset \cite{C16} for 60 epochs using a fixed learning rate of 0.0016. After that, we further fine-tuned the model for another 40 epochs with a linear schedule learning rate at base  of 0.0008. Data argumentation including random scaling, crops and flips were used to further improve the robustness of the model \cite{C17,C18,C19}. The training image batch size was set to 10 and the training process took approximately 7 days on two 12GB Titan X GPUs (Maxwell architecture).   

\par

During segmentation of the test set, we first use the ResNet to produce the probability map of $Y^t$ and $Y^v$ based on the input image $I$. After that, we used the cascaded ResNet together with the input image $I$ and the two probability map $Y^t$, $Y^v$ to generate the final prediction. The prediction time was approximately 48 minutes (multi-scale) or 16 minutes (single-scale) for CT volume with an average of 390 axial slices.

\section{Experiments and Results}
\label{sec:Experiments and Results}

\subsection{Experimental Setup}
\label{sssec:Experimental Setup}

As the test ground truth was not available at the time this manuscript was prepared, we provide the evaluation conducted on the validation dataset (13 studies); the Dice and Jaccard indices were used to measure the segmentation accuracy. We compared our method with: (i) traditional FCN model based on VGGNet architecture; (ii) the ResNet architecture; (iii) our cascaded ResNet; (iv) cascaded ResNet with 3D conditional random field (3D-CRF \cite{C20}) as a posting processing approach; and (v) cascaded ResNet with multi-scale fusion.

\subsection{Results }
\label{sssec:Results }

Table \ref{table:tab2} shows the liver and liver lesion segmentation results. The cascaded ResNet with multi-scale fusion achieved the best segmentation results. This method also had a higher accuracy than traditional VGGNet based FCN: an improvement of 3.94\% in the Dice index for liver segmentation and 20.13\% for liver lesion segmentation.

\begin{figure*}
	\begin{center}
			\includegraphics[width=0.8\textwidth]{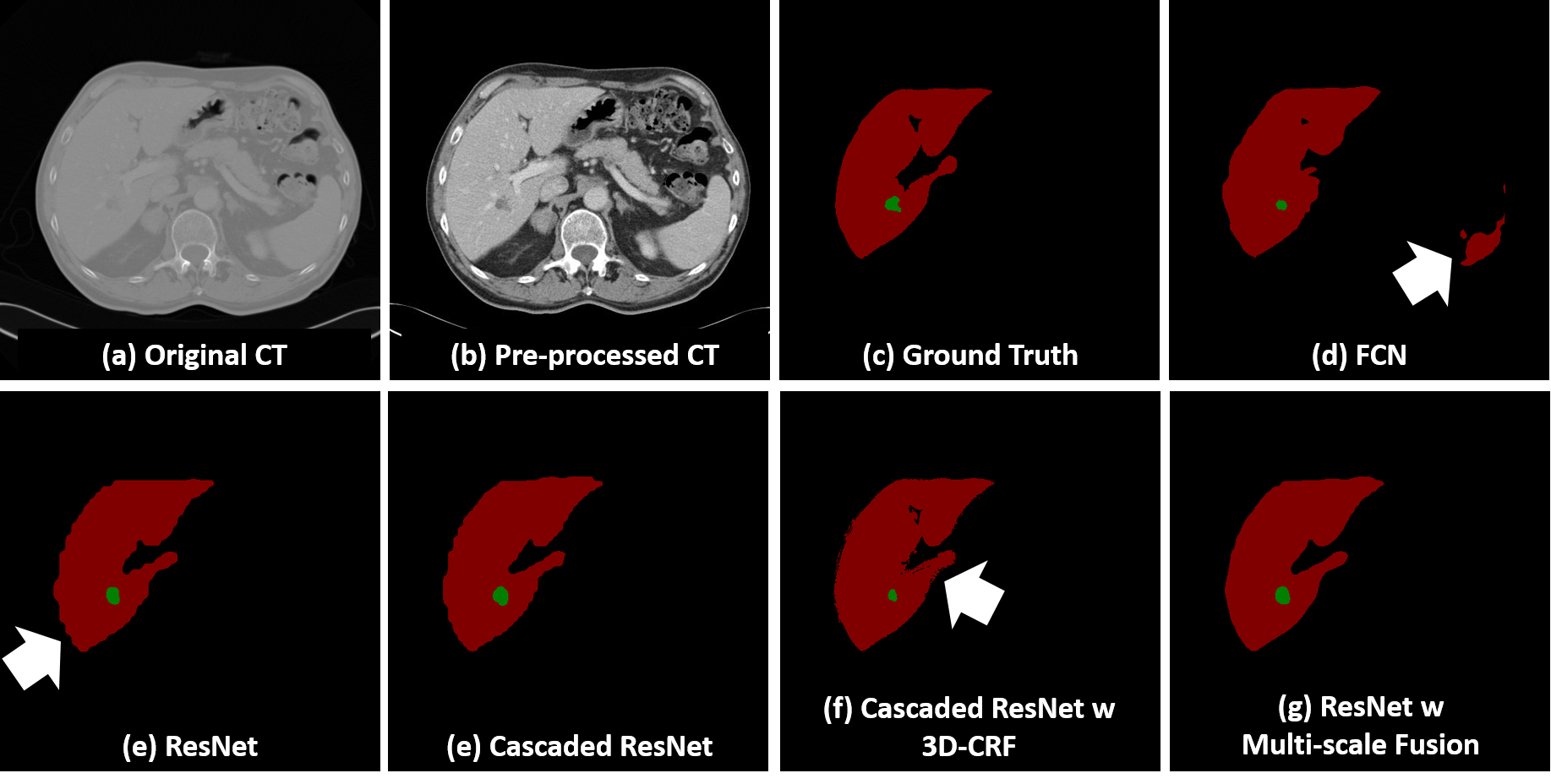}
	\end{center}
	\caption{Liver and liver lesion segmentation results from one example study.}
	\label{fig:3}
\end{figure*}

\begin{table*}[t]
\centering
\caption{Comparison of the liver and liver lesion segmentation for different methods.}
\label{table:tab2}
\begin{tabular}{ccccc}
\hline
\textbf{\%}                                     & \textbf{Liver Dice} & \textbf{Lesion Dice} & \textbf{Liver Jaccard} & \textbf{Lesion Jaccard} \\ \hline
\textbf{VGG-FCN}                                & 91.96               & 29.88                & 85.47                  & 19.65                   \\ \hline
\textbf{ResNet}                                 & 95.27               & 48.50                 & 91.17                  & 37.24                   \\ \hline
\textbf{Cascaded ResNet}                        & 95.51               & 49.83                & 91.45                  & 38.59                   \\ \hline
\textbf{Cascaded ResNet (w 3D-CRF)}             & 95.24               & 31.65                & 91.01                  & 23.86                   \\ \hline
\textbf{Cascaded ResNet (w Multi-scale Fusion)} & \textbf{95.90}       & \textbf{50.01}       & \textbf{92.19}         & \textbf{38.79}          \\ \hline
\end{tabular}
\end{table*}

\section{Discussions}
\label{sec:Discussions}

The difference between the VGG-based FCN and other methods demonstrate the advantages of the deep residual architecture for segmentation. The proposed cascaded ResNet further improved the segmentation results, especially for the liver lesion segmentation (1.33\% improvement in the Dice index). We attribute this ability of the cascaded architecture to iteratively refine the segmentation results of both liver and liver lesions using high-level of semantic differences between these structures, as opposed to a reliance on low-level pixel values, which can be shared. The 3D-CRF model had a reduced liver lesion segmentation accuracy when compared to the base ResNet model on both the original volumes and the isotropically rescaled volumes. We attribute this reduced performance to the reliance of CRF on low-level features, which are incapable of separating the liver lesions from the surrounding liver tissues (Fig. \ref{fig:3}). The cascaded ResNet with multi-scale fusion achieved the best results. This is due to the fact that the CT studies derived from different medical centers varies in staging and pixel resolution. The multi-scale fusion approach is scale-invariant and therefore produced the best results overall. For this reason, in our submission to the challenge, we used the cascaded ResNet with multi-scale fusion method on the test dataset; we achieved the 4\textsuperscript{th} on the online leaderboard\footnote{https://competitions.codalab.org/competitions/15595\#results} by the submission deadline with an overall Dice index of 64.00\% for the liver lesion segmentation.

\bibliographystyle{IEEEbib}
\bibliography{strings,refs}

\end{document}